\documentclass{article}

\usepackage{microtype}
\usepackage{graphicx}
\usepackage{subcaption}
\usepackage{booktabs}
\usepackage{hyperref}

\usepackage[accepted]{icml2026}

\usepackage{amsmath}
\usepackage{amssymb}
\usepackage{mathtools}
\usepackage{amsthm}
\usepackage[capitalize,noabbrev]{cleveref}
\usepackage{xurl}

\theoremstyle{plain}

\theoremstyle{definition}

\theoremstyle{remark}

\icmltitlerunning{MARL from Delayed Marketplace Feedback for Objective-Weight Adaptation in Three-Sided Dispatch}

\begin{document}

\twocolumn[
  \icmltitle{Multi-Agent Reinforcement Learning from Delayed Marketplace Feedback for Objective-Weight Adaptation in Three-Sided Dispatch}

  % Replace the placeholder author block below for non-blind versions.
  
  % \begin{icmlauthorlist}
  %   \icmlauthor{Anonymous Authors}{anon}
  % \end{icmlauthorlist}

  % \icmlaffiliation{anon}{Anonymous Institution}
  % \icmlcorrespondingauthor{Anonymous Authors}{anonymous@anonymous.edu}

    \begin{icmlauthorlist}
    \icmlauthor{Haochen Wu}{1}
    \icmlauthor{Yi Hou}{1}
    \icmlauthor{Ryan Xie}{1}
  \end{icmlauthorlist}

  \icmlaffiliation{1}{DoorDash Inc}
  \icmlcorrespondingauthor{Haochen Wu}{haochen.wu@doordash.com}

  \icmlkeywords{offline reinforcement learning, multi-agent reinforcement learning, multi-objective decision making, marketplace dispatch, switchback experiments}

  \vskip 0.3in
]

\printAffiliationsAndNotice{}

\begin{abstract}
Dispatch in three-sided marketplaces provides a natural setting for reinforcement learning from world feedback: decisions are evaluated by delayed operational outcomes such as delivery speed, courier utilization, and merchant congestion. We present a deployed reinforcement learning system at DoorDash that adapts dispatch objective weights in a large-scale food-delivery marketplace using delayed signals. Rather than replacing the combinatorial assignment optimizer, a store-level policy learned from logged marketplace data selects a discrete multiplier that shifts the dispatch optimizer’s tradeoff between delivery quality and batching efficiency. This interface enables offline policy learning under noisy, delayed, and coupled feedback while preserving production feasibility constraints and operational safeguards. We train a shared value function using centralized offline data and decentralized store-level execution, with Double Q-learning targets and a conservative regularizer to reduce out-of-distribution value overestimation. In a production switchback experiment, the offline-trained policy increases batching and reduces courier-side time costs without degrading customer-facing delivery quality. Results illustrate how world feedback from a live economic and logistics system can be used to safely adapt decision policies online.
\end{abstract}

\section{Introduction}
Food-delivery dispatch is a sequential decision problem embedded in a three-sided marketplace. Each decision affects customer delivery quality, merchant congestion, courier availability, batching opportunities, and cost effectiveness, which are evaluated not by human preference labels but by delayed operational outcomes from the marketplace. Dispatch systems must therefore balance competing objectives: batching can improve courier efficiency, while faster execution can reduce lateness and improve customer experience \citep{ulmer2021restaurant,agatz2023crowdsourced}. In production, this tradeoff is often governed by static heuristic weights that are tuned globally and updated manually, making them brittle under local, time-varying conditions: over-batching during congestion can increase lateness, while under-batching during slack periods leaves efficiency gains unrealized.

\begin{figure}[t]
    \centering
    \includegraphics[width=\linewidth]{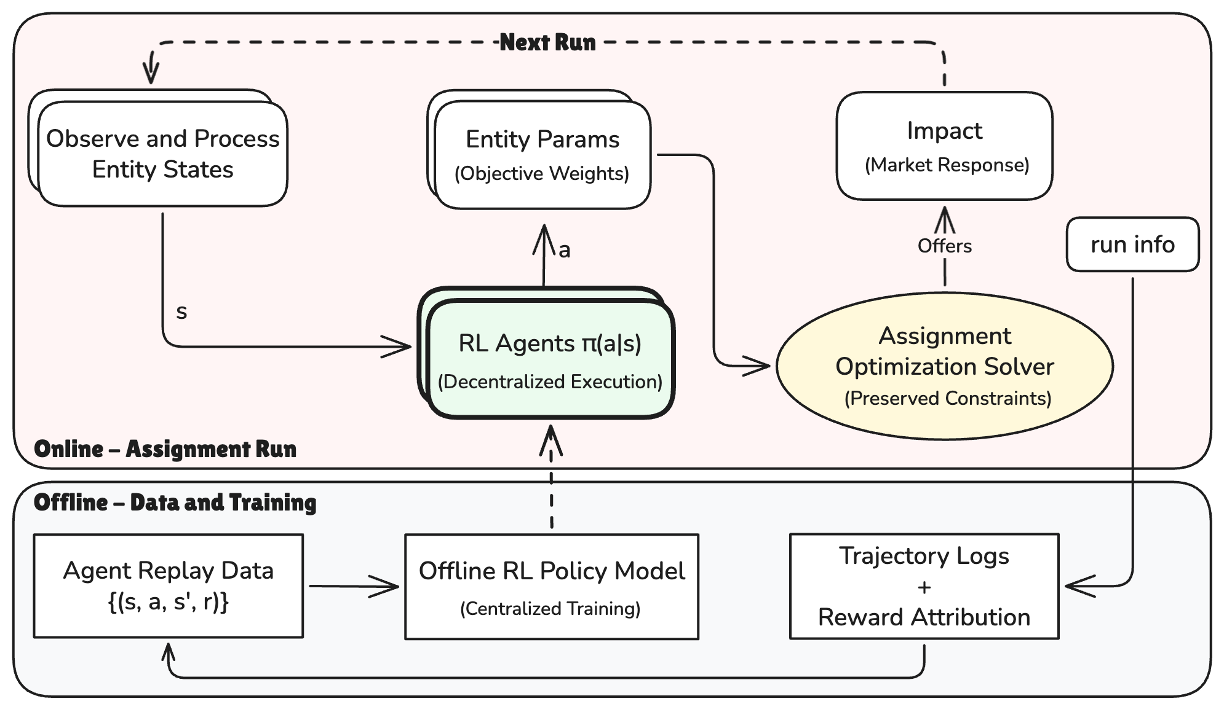}
    % \caption{
    % End-to-end workflow for dispatch objective-weight adaptation. Online, a store-level policy maps local state to an objective-weight multiplier consumed by the assignment optimizer. Offline, logged decisions are joined with delayed marketplace outcomes to construct rewards and train the next policy.
    % }
    \caption{
Agentic objective-weight adaptation loop for production dispatch. 
During online serving, a policy agent observes local states and selects an objective-weight multiplier that parameterizes the assignment optimizer. The optimizer remains responsible for feasible courier-order assignment decisions. During offline learning, logged runs are joined with delayed marketplace outcomes to construct transition data tuples for policy training and deployment.
}
    \label{fig:rl-dispatch}
\end{figure}

\begin{figure*}[t]
    \centering
    \includegraphics[width=0.92\textwidth]{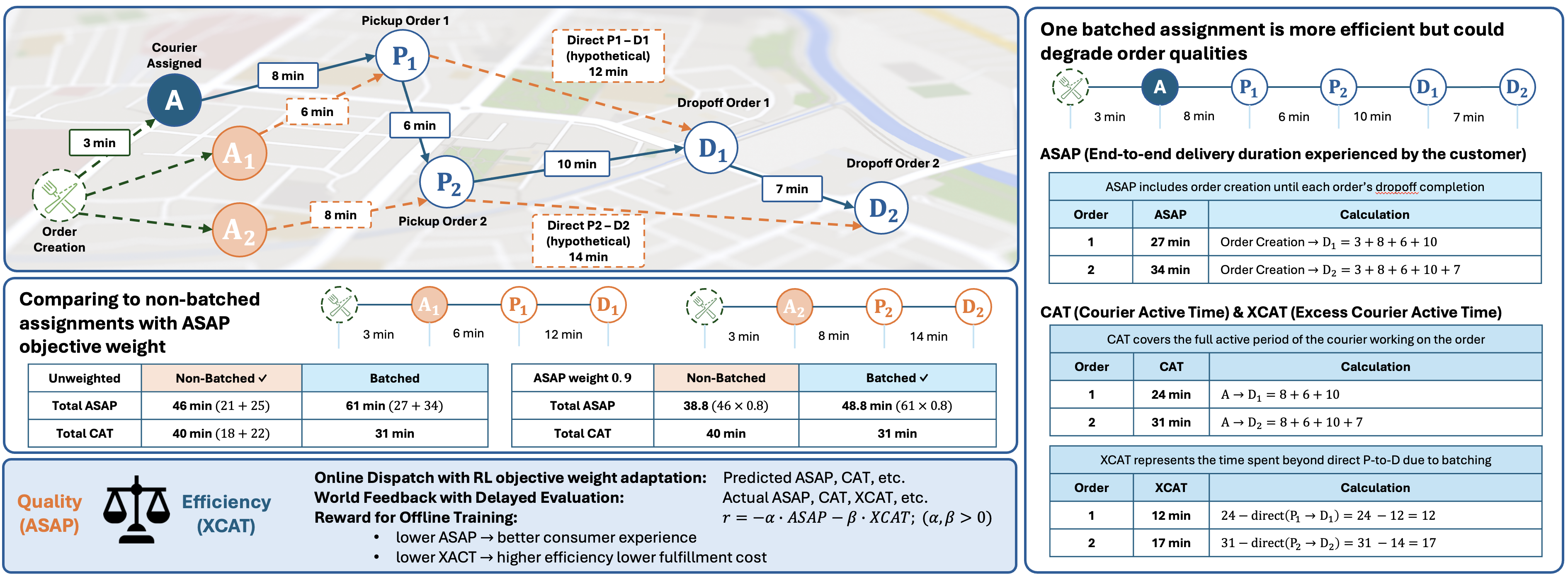}
    % \caption{Logistics timing components. $\mathrm{ASAP}$ measures customer-facing delivery duration, $\mathrm{CAT}$ measures courier active time, and $\mathrm{XCAT}$ measures excess active time. The RL-selected ASAP weight steers the optimizer's speed-efficiency tradeoff. The delayed ASAP and XCAT real-world signals are joined back to each run for reward shaping.
    % }
    \caption{
Logistics timing compoents used for delayed reward attribution. 
ASAP measures customer-facing delivery duration, and XCAT measures excess courier active time (CAT) beyond the direct route. 
The RL-selected ASAP-weight multiplier steers the optimizer's speed-efficiency tradeoff: higher weights favor faster delivery completion, while lower weights make batching-compatible assignments more attractive. 
Observed ASAP and XCAT outcomes are joined back to dispatch runs for reward construction.
% Logistics timing signals for delayed reward attribution. ASAP captures customer-facing delivery duration, while XCAT captures excess courier active time beyond the direct route. The RL-selected ASAP-weight multiplier steers the optimizer’s speed-efficiency tradeoff, and observed ASAP/XCAT outcomes are joined back to dispatch runs for reward construction.
}
    \label{fig:rl-reward}
\end{figure*}

We present a deployed offline reinforcement learning (RL) system for real-time objective-weight adaptation in DoorDash’s large-scale food-delivery dispatch platform. Rather than replacing the combinatorial assignment optimizer, the learned policy controls a narrow objective-weight interface that shifts the optimizer's tradeoff between delivery speed and batching efficiency. Every dispatch run, each store-level RL agent observes local marketplace state and selects a discrete multiplier applied to the optimizer objective. This constrained action space preserves existing optimization and operational safeguards while enabling local adaptation. The learned policy is trained offline from logged marketplace outcomes and deployed at production scale, serving hundreds of millions of daily inferences at a 20-second cadence.

Prior work has applied RL to food-delivery operations, including dispatching, courier assignment, routing, repositioning, and batching \citep{chen2024matching,jahanshahi2022deep,zou2021online,guo2021concurrent,lu2024abod,cheng2025realtime}. Most methods learn direct operational decisions and are evaluated in simulated or offline settings. We study a complementary production setting where the RL agent learns from delayed marketplace feedback while modulating the objective of an existing assignment optimizer through a constrained adaptation interface. This setting introduces delayed and coupled feedback, as dispatch decisions jointly affect quality for customers, merchant congestion, and courier utilization.

Our contributions are threefold:
1) Introducing an RL architecture that learns from marketplace feedback while adapting dispatch objective weights through a low-dimensional control interface rather than replacing the production assignment optimizer.
2) Formulating objective-weight adaptation as an offline multi-agent decision-making problem with store-level decentralized execution and delayed regional rewards from the real-world marketplace.
3) Providing production switchback evidence showing increased batching and reduced courier-side time costs without degrading customer-facing delivery quality.

\section{Learning from Marketplace Feedback}
The deployed system has two nested decision layers. The inner is the optimizer, mapping orders, couriers, constraints, and objective weights to courier-order assignments. The outer is the objective-weight adaptation agent (OWA-RL) that learns from marketplace outcomes and selects a store-level objective-weight multiplier before optimization.

\paragraph{Formulation}
We formulate this outer adaptation problem as a decentralized multi-agent Markov decision process with centralized offline training. Let $\mathcal{I}=\{1,\ldots,N\}$ denote stores, each treated as an agent. At assignment cycle $t$, store $i$ observes local state $s_t^i\in\mathcal{S}$ and selects action $a_t^i\in\mathcal{A}$. The action changes the downstream optimizer objective and the resulting assignment decisions induce delayed marketplace outcomes. We write the RL problem as $\mathcal{M}=(\mathcal{I},\mathcal{S},\mathcal{A},\mathcal{T},R,\gamma,\lambda_0)$, where $\mathcal{T}$ denotes transition dynamics induced jointly by marketplace evolution and the production optimizer, $R$ is a delayed multi-objective reward, $\gamma$ is the discount factor, and $\lambda_0$ is the baseline delivery-speed objective weight (ASAP weight).

\paragraph{Action}
$\mathcal{A}=\{0.8,0.9,1.0,1.1,1.2\}$. The action is a discrete multiplier on the baseline ASAP weight. Store $i$ selects $a_t^i \in \mathcal{A}$, producing the adapted weight $\lambda_t^i=a_t^i\lambda_0$ for the current assignment cycle. Lower weights make batching or more efficient routing more attractive during optimization (the solid line in Figure~\ref{fig:rl-reward}). Higher weights emphasize faster order completion and may lead the optimizer to assign nearby orders to separate couriers rather than batch them (the dashed lines in Figure~\ref{fig:rl-reward}). The neutral action $a_t^i=1.0$ recovers the static production baseline. Thus, the policy steers the existing optimizer through a constrained objective-weight interface, while the optimizer enforces feasibility and makes the final courier-order assignments.

\paragraph{State}
$s_t^i =
    \left[
    d_t^i,\,
    \mathrm{sup}_t^i,\,
    \mathrm{cwt}_t^i
    \right]$.
The store-level state is refreshed every assignment run. Here, $d_t^i$ is outstanding delivery count, $\mathrm{cwt}_t^i$ is median courier wait time, and $\mathrm{sup}_t^i$ is a localized supply-pressure feature. To expose store-level variation hidden by regional supply signals, we rescale the regional feature by effective courier supply as $\mathrm{sup}_t^i=\mathrm{sup}_t^{g(i)}\cdot\frac{\widetilde{S}_t^{g(i)}}{\widetilde{S}_t^i}$,
where $g(i)$ is the region containing store $i$, $\widetilde{S}_t^i$ is the recent median number of feasible couriers reaching optimization for store $i$, and $\widetilde{S}_t^{g(i)}$ is the corresponding regional median. When the store has more feasible couriers than the regional median (${\widetilde{S}_t^{g(i)}}>{\widetilde{S}_t^i}$), it results in lower supply pressure.

\paragraph{Reward.}
Rewards are computed from delayed delivery outcomes and aggregated regionally to capture the network effect across nearby stores and couriers. Figure~\ref{fig:rl-reward} summarizes the timing quantities used in the reward. For delivery $j$, with order creation, courier acceptance, and dropoff times $t_j^{\mathrm{create}}$, $t_j^{\mathrm{acc}}$, and $t_j^{\mathrm{drop}}$, define
\begin{equation}
    \mathrm{ASAP}_j
    =
    t_j^{\mathrm{drop}} - t_j^{\mathrm{create}},
    \qquad
    \mathrm{CAT}_j
    =
    t_j^{\mathrm{drop}} - t_j^{\mathrm{acc}},
\end{equation}
\begin{equation}
    \mathrm{XCAT}_j
    =
    \mathrm{CAT}_j
    -
    T_j^{\mathrm{direct}},
\end{equation}
where $T_j^{\mathrm{direct}}$ is direct pickup-to-dropoff travel time if delivery $j$ were served alone. Let $\mathcal{D}_t^{g(i)}$ be deliveries in region $g(i)$ whose outcomes are attributed to decision cycle $t$. The reward is
\begin{equation}
    r_t^{g(i)}
    =
    -
    \frac{1}{|\mathcal{D}_t^{g(i)}|}
    \sum_{j \in \mathcal{D}_t^{g(i)}}
    \left(
        \alpha \, \mathrm{ASAP}_j
        +
        \beta \, \mathrm{XCAT}_j
    \right).
\end{equation}
The reward is a compact representation of the marketplace feedback. ASAP captures customer-facing latency, XCAT captures excess courier-side effort beyond direct travel, and regional aggregation captures network effects across nearby stores and couriers.

\paragraph{Offline Training}

Let $\pi_\theta(a\mid s)$ denote the shared store-level policy. Stores execute the policy independently using local state, while training pools experience across stores and uses regional rewards to capture cross-store network effects. A transition pipeline joins logged online decisions with delayed fulfillment outcomes to construct $\mathcal{D}=\{(s_t^i,a_t^i,r_t^{g(i)},s_{t+1}^i)\}$, where $g(i)$ is the dispatch region containing store $i$ and $r_t^{g(i)}$ is the realized regional reward attributed to decision cycle $t$. The objective is to learn a policy maximizing expected discounted regional reward, $\pi^*=\arg\max_{\pi_\theta}\mathbb{E}_{\pi_\theta}\left[\sum_{k=0}^{H}\gamma^k r_{t+k}^{g(i)}\right]$.

We train a discrete-action value function $Q_\theta(s,a)$ using an offline Double DQN objective \citep{mnih2015dqn,vanhasselt2016double}, where the behavior network selects the next action and the target network evaluates it with temporal-difference loss:
\begin{equation}
    y_t^{\mathrm{DDQN}}
    =
    r_t^{g(i)}
    +
    \gamma
    Q_{\bar{\theta}}
    \left(
        s_{t+1}^i,
        \arg\max_{a' \in \mathcal{A}}
        Q_\theta(s_{t+1}^i,a')
    \right).
\end{equation}
\begin{equation}
    \mathcal{L}_{\mathrm{DDQN}}(\theta)
    =
    \mathbb{E}_{(s,a,r,s') \sim \mathcal{D}}
    \left[
    \left(
    Q_\theta(s,a) - y^{\mathrm{DDQN}}
    \right)^2
    \right].
\end{equation}

Because the learned policy may assign high values to actions weakly supported by logged data, we add a discrete Conservative Q-Learning regularizer \citep{kumar2020cql}:
\begin{equation}
    \mathcal{L}_{\mathrm{CQL}}(\theta)
    =
    \mathbb{E}_{(s,a) \sim \mathcal{D}}
    \left[
    \log \sum_{a' \in \mathcal{A}}
    \exp Q_\theta(s,a')
    -
    Q_\theta(s,a)
    \right].
\end{equation}
The final training objective is
\begin{equation}
    \mathcal{L}(\theta)
    =
    \mathcal{L}_{\mathrm{DDQN}}(\theta)
    +
    \eta
    \mathcal{L}_{\mathrm{CQL}}(\theta),
\end{equation}
where $\eta$ controls the strength of conservatism. Double DQN reduces maximization bias, while the conservative penalty discourages unsupported actions from receiving high values, improving offline training stability before deployment.

Appendix~\ref{app:training} details the controlled offline data collection, policy architecture,
and training hyperparameters. Appendix~\ref{app:training-curves} compares training curves with a DQN
baseline, showing how the conservative regularizer affects offline learning before
online deployment.

\section{Production Switchback Experiment}

\begin{table*}[t]
\centering
\caption{Online experiment results for all day parts and dinner. Baseline and OWA-RL report metric means, while ATE reports the variance-reduction adjusted average treatment effect, with $p$-values shown in parentheses. CAT, CWT, and ASAP are measured in seconds; \% batched and \% 20-min late are reported as percentage-point rates. Lower values are better for CAT, CWT, ASAP, and \% 20-min late, while higher values are better for \% batched. Statistically significant ATEs at $p<0.05$ are bolded.}
\label{tab:online_results}
\small
\setlength{\tabcolsep}{5pt}
\begin{tabular}{ll|ccc|cc}
\toprule
&
& \multicolumn{3}{c|}{Efficiency Metrics}
& \multicolumn{2}{c}{Quality Metrics} \\
\cmidrule(lr){3-5}\cmidrule(lr){6-7}
Scope
& Statistic
& CAT (sec.)
& CWT (sec.)
& \% batched
& ASAP (sec.)
& \% 20-min late \\
\midrule
All Day Parts
& Baseline
& $1163.0$
& $277.1$
& $47.52\%$
& $1956.0$
& $2.09\%$ \\
& OWA-RL
& $1159.8$
& $275.7$
& $48.14\%$
& $1960.0$
& $2.09\%$ \\
& ATE $(p)$
& $\mathbf{-1.261}$ $(\mathbf{0.019})$
& $\mathbf{-0.856}$ $(\mathbf{0.004})$
& $\mathbf{+0.495}$ $(\mathbf{<0.001})$
& $+0.972$ $(0.264)$
& $-0.012$ $(0.237)$ \\
\midrule
Dinner
& Baseline
& $1156.3$
& $262.9$
& $57.99\%$
& $2168.3$
& $2.36\%$ \\
& OWA-RL
& $1153.1$
& $261.6$
& $58.59\%$
& $2173.0$
& $2.34\%$ \\
& ATE $(p)$
& $\mathbf{-1.289}$ $(\mathbf{0.042})$
& $\mathbf{-1.030}$ $(\mathbf{0.041})$
& $\mathbf{+0.600}$ $(\mathbf{0.010})$
& $+0.869$ $(0.633)$
& $\mathbf{-0.037}$ $(\mathbf{0.040})$ \\
\bottomrule
\end{tabular}
\label{tab:impact}
\end{table*}

\paragraph{Experiment Setup}

Before deployment, we use offline reward-reweighting diagnostics to test whether the learned policy responds directionally to different definitions of marketplace feedback. We choose $\alpha=0.9$ because it yields a more balanced action distribution, avoiding collapse toward either aggressive batching or speed prioritization. Full diagnostics are provided in Appendix~\ref{app:reward-reweighting}.

We evaluate OWA-RL against the DoorDash production baseline using a global switchback experiment. The randomization units are approximately 4,000 geographic regions. At each two-hour switchback interval, regions are randomly assigned to treatment or control, with approximately half of regions exposed to each condition in each interval, over a two-week experiment period. The control uses the baseline static objective weight, while the treatment uses the OWA-RL policy trained with $\alpha=0.9$. Eligible traffic includes all stores in randomized regions.

Treatment changes only weight selection; the optimizer, constraints, and serving infrastructure remain fixed. We estimate treatment effects using CUPED variance reduction and compute p-values clustered at the region-hour switchback bucket. Following standard switchback-experiment practice, we monitor customer-experience quality, cancellations, and carryover effects as guardrails~\citep{bojinov2022switchback}.

\paragraph{Policy Behavior}

\begin{figure}[t]
    \centering
    \includegraphics[width=\linewidth]{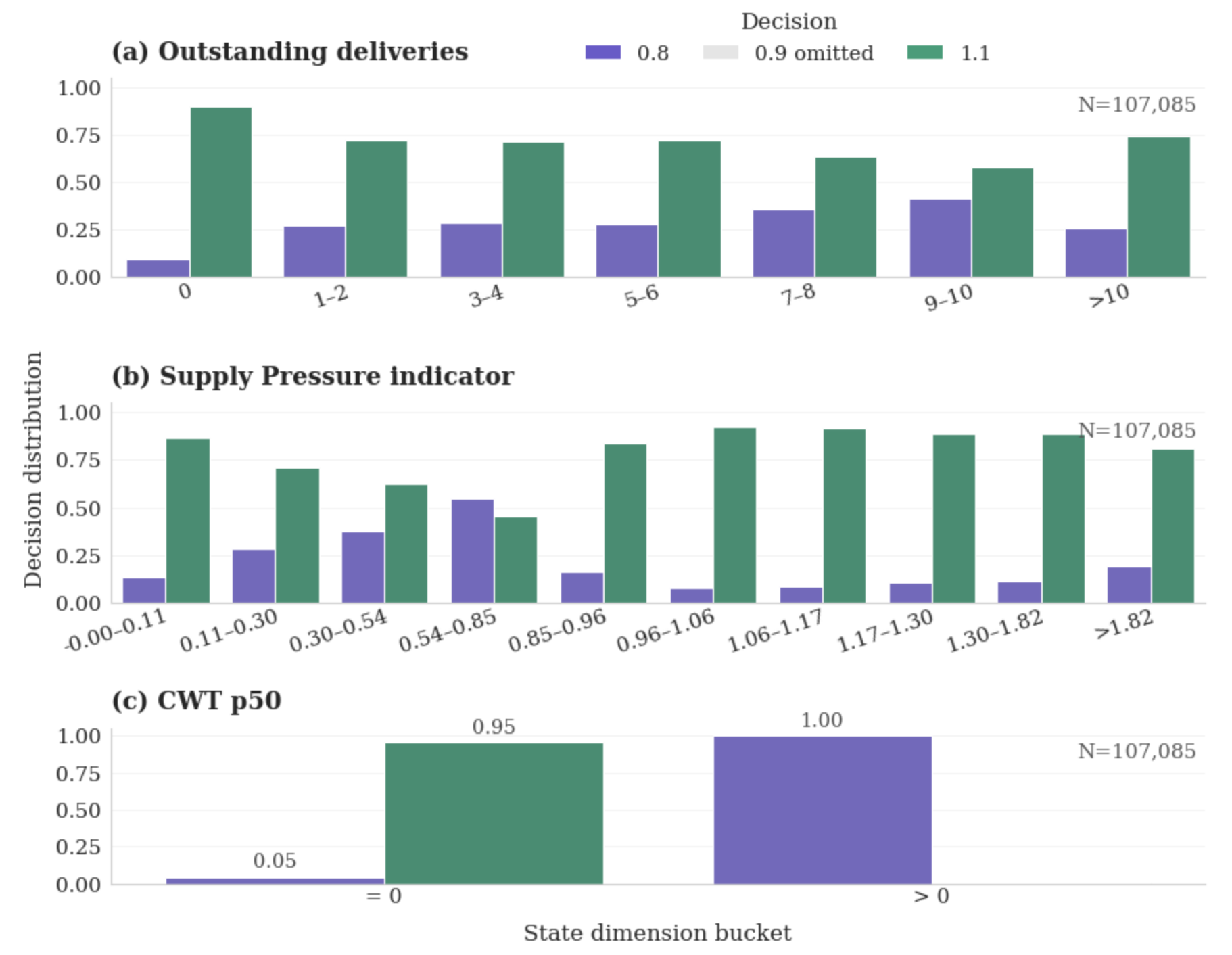}
    % \caption{
    % Empirical action distribution across state dimensions, computed from SF Bay market inference logs on Friday 4/25 during local hour 5--8pm. The policy varies ASAP-weight multipliers across backlog, supply-pressure, and courier-wait-time buckets.
    % }
    \caption{
Empirical action distributions from DoorDash production logs during a Friday dinner peak, showing state-dependent ASAP-weight multipliers across backlog, supply pressure, and CWT.
% Empirical action distributions across state dimensions from DoorDash production inference logs in the San Francisco Bay Area during a Friday dinner peak. The policy varies ASAP-weight multipliers across backlog, supply-pressure, and CWT buckets.
}
    \label{fig:empirical-actions}
\end{figure}

To verify the deployed policy uses local marketplace state, we inspect predicted actions from the San Francisco-Bay Area market during Friday dinner peak. Figure~\ref{fig:empirical-actions} shows that the policy shifts probability mass across ASAP-weight multipliers as outstanding deliveries, supply pressure, and courier wait time change. The dominant action is often the higher ASAP-weight multiplier, but the lower multiplier becomes more likely in specific backlog and supply regimes. We monitor how decisions and supply pressure states change week over week in Appendix~\ref{app:drift-monitoring}, indicating state-dependent adaptation rather than fixed global retuning.

\paragraph{Online Impact}
Table~\ref{tab:impact} reports how a policy trained from logged world feedback transfers to online marketplace operation and summarizes results for all day parts and dinner, including baseline and OWA-RL means, CUPED-adjusted average treatment effects (ATE), and clustered p-values. Across all day parts, OWA-RL improves efficiency without degrading delivery quality: CAT and CWT decrease significantly, batching increases by 0.495 percentage points, and ASAP and 20-minute lateness remain statistically unchanged. CWT is important because courier wait time at merchant pickup contributes to courier utilization and indicates non-optimal arrival timing relative to order readiness; it complements CAT, which primarily captures active travel and service time. During dinner, the policy again reduces CAT and CWT and increases batching, while ASAP remains unchanged, and 20-minute lateness improves slightly. These results show that adaptive objective-weight selection increases batching and reduces courier-side time costs without degrading customer-facing quality at production scale.

\section{Conclusion}
We presented a production case study of reinforcement learning from delayed marketplace feedback for real-time objective-weight adaptation in a three-sided dispatch system at DoorDash. By exposing a decision layer before the existing combinatorial assignment optimizer, online experiments show the learned policy increases batching, reduces courier-side time costs, preserves global delivery quality and operational safeguards, and improves dinner-hour lateness.

The current approach only acts through a low-dimensional objective-weight interface, which ensures serving reliability, scalability, and latency. Rewards are attributed using delayed regional outcomes, better reflecting the marketplace network effect, but introducing noisy credit assignment for individual store-level actions. Also, the reliability of the offline-trained policy heavily depends on the stability of marketplace dynamics after deployment.

Future work includes introducing different decision layers to the dispatch system. This would require studying systematic methods for detecting distribution shift and RL decision interactions in the multi-agent setting. We have extended the monitoring of state, action, and reward distributions, with examples shown in Appendix~\ref{app:drift-monitoring}. In addition, interpretability tools based on large-language-models that link RL policy decisions to dynamic states could support automated hypothesis generation, debugging, and policy retraining.

\newpage
\section*{Impact Statement}
This paper presents work with the goal to advance reinforcement learning methods for large-scale logistics and marketplace dispatch. Since dispatch decisions affect customers, merchants, and couriers, production use should include monitoring of service-quality guardrails, courier wait time, and workload effects, regional heterogeneity, distribution shift, and rollback criteria. This work also emphasizes the importance of having a constrained control layer, offline RL safeguards, and online experimentation when deploying reinforcement learning in systems with real-world operational and labor impacts.

\bibliography{references}

@article{ulmer2021restaurant,
  title={The Restaurant Meal Delivery Problem: Dynamic Pickup and Delivery with Deadlines and Random Ready Times},
  author={Ulmer, Marlin W. and Thomas, Barrett W. and Campbell, Ann M. and Woyak, Nicholas},
  journal={Transportation Science},
  volume={55},
  number={1},
  pages={75--100},
  year={2021},
  doi={10.1287/trsc.2020.1000}
}

@article{agatz2023crowdsourced,
  title={Crowdsourced on-demand food delivery: An order batching and assignment algorithm},
  author={Agatz, Niels and Fan, Yanjun and Stam, Daan},
  journal={Transportation Research Part C: Emerging Technologies},
  volume={149},
  pages={104055},
  year={2023},
  doi={10.1016/j.trc.2023.104055}
}

@article{chen2024matching,
  title={A Matching Algorithm with Reinforcement Learning and Decoupling Strategy for Order Dispatching in On-Demand Food Delivery},
  author={Chen, Jingfang and Wang, Ling and Pan, Zixiao and Wu, Yuting and Zheng, Jie and Ding, Xuetao},
  journal={Tsinghua Science and Technology},
  volume={29},
  number={2},
  pages={386--399},
  year={2024},
  doi={10.26599/TST.2023.9010069}
}

@article{jahanshahi2022deep,
  title={A Deep Reinforcement Learning Approach for the Meal Delivery Problem},
  author={Jahanshahi, Hadi and Bozanta, Aysun and Cevik, Mucahit and Kavuk, Eray Mert and Tosun, Ay{\c{s}}e and Sonuc, Sibel B. and Kosucu, Bilgin and Ba{\c{s}}ar, Ay{\c{s}}e},
  journal={Knowledge-Based Systems},
  volume={243},
  pages={108489},
  year={2022},
  doi={10.1016/j.knosys.2022.108489}
}

@article{zou2021online,
  title={Online Food Ordering Delivery Strategies Based on Deep Reinforcement Learning},
  author={Zou, Guangyu and Tang, Jiafu and Yilmaz, Levent and Kong, Xiangyu},
  journal={Applied Intelligence},
  year={2021},
  doi={10.1007/s10489-021-02750-3}
}

@inproceedings{guo2021concurrent,
  author    = {Baoshen Guo and Shuai Wang and Yi Ding and Guang Wang and Suining He and Desheng Zhang and Tian He},
  title     = {Concurrent Order Dispatch for Instant Delivery with Time-Constrained Actor-Critic Reinforcement Learning},
  booktitle = {Proceedings of the IEEE Real-Time Systems Symposium (RTSS)},
  year      = {2021}
}

@article{lu2024abod,
  author  = {Miaojia Lu and Xinyu Yan and Shadi Sharif Azadeh and Pengling Wang},
  title   = {An Adaptive Agent-Based Approach for Instant Delivery Order Dispatching: Incorporating Task Buffering and Dynamic Batching Strategies},
  journal = {International Journal of Transportation Science and Technology},
  volume  = {13},
  pages   = {137--154},
  year    = {2024},
  doi     = {10.1016/j.ijtst.2023.12.006}
}

@misc{cheng2025realtime,
  author       = {Jingyi Cheng and Shadi Sharif Azadeh},
  title        = {Real-Time Integrated Dispatching and Idle Fleet Steering with Deep Reinforcement Learning for a Meal Delivery Platform},
  year         = {2025},
  eprint       = {2501.05808},
  archivePrefix= {arXiv},
  primaryClass = {cs.LG},
  url          = {https://arxiv.org/abs/2501.05808}
}

@article{mnih2015dqn,
  author  = {Volodymyr Mnih and Koray Kavukcuoglu and David Silver and Andrei A. Rusu and Joel Veness and Marc G. Bellemare and Alex Graves and Martin Riedmiller and Andreas K. Fidjeland and Georg Ostrovski and Stig Petersen and Charles Beattie and Amir Sadik and Ioannis Antonoglou and Helen King and Dharshan Kumaran and Daan Wierstra and Shane Legg and Demis Hassabis},
  title   = {Human-Level Control through Deep Reinforcement Learning},
  journal = {Nature},
  volume  = {518},
  number  = {7540},
  pages   = {529--533},
  year    = {2015},
  doi     = {10.1038/nature14236}
}

@inproceedings{vanhasselt2016double,
  author    = {Hado van Hasselt and Arthur Guez and David Silver},
  title     = {Deep Reinforcement Learning with Double Q-Learning},
  booktitle = {Proceedings of the AAAI Conference on Artificial Intelligence},
  volume    = {30},
  number    = {1},
  year      = {2016}
}

@inproceedings{kumar2020cql,
  author    = {Aviral Kumar and Aurick Zhou and George Tucker and Sergey Levine},
  title     = {Conservative Q-Learning for Offline Reinforcement Learning},
  booktitle = {Advances in Neural Information Processing Systems},
  volume    = {33},
  pages     = {1179--1191},
  year      = {2020}
}

@article{bojinov2022switchback,
  title={Design and Analysis of Switchback Experiments},
  author={Bojinov, Iavor and Simchi-Levi, David and Zhao, Jinglong},
  journal={Management Science},
  volume={69},
  number={7},
  pages={3759--3777},
  year={2023},
  doi={10.1287/mnsc.2022.4583}
}
\bibliographystyle{icml2026}

%%%%%%%%%%%%%%%%%%%%%%%%%%%%%%%%%%%%%%%%%%%%%%%%%%%%%%%%%%%%%%%%%%%%%%%%%%%%%%%
%%%%%%%%%%%%%%%%%%%%%%%%%%%%%%%%%%%%%%%%%%%%%%%%%%%%%%%%%%%%%%%%%%%%%%%%%%%%%%%
% APPENDIX
%%%%%%%%%%%%%%%%%%%%%%%%%%%%%%%%%%%%%%%%%%%%%%%%%%%%%%%%%%%%%%%%%%%%%%%%%%%%%%%
%%%%%%%%%%%%%%%%%%%%%%%%%%%%%%%%%%%%%%%%%%%%%%%%%%%%%%%%%%%%%%%%%%%%%%%%%%%%%%%
\newpage
\appendix
\onecolumn
\section{Offline Training}
\label{app:training}

\subsection{Data Collection}
Training data is collected over two iterations using a controlled regional rollout. 
In each iteration, data collection lasts approximately two days. To limit potential 
user-facing impact, we randomly select only $0.5\%$ of global regions every two hours 
for data collection. This rollout design keeps the intervention localized while 
allowing us to measure cumulative impact within each two-hour window.

In the first iteration, data is collected using pure exploration, where actions are 
sampled uniformly at random from the action space. After training an initial policy 
on this dataset, we conduct a second data-collection iteration using a mixture policy 
with $50\%$ exploitation and $50\%$ exploration. Specifically, with probability $0.5$, 
the learned policy is used to select the action, and with probability $0.5$, a random 
action is sampled from the action space.

Each logged transition is represented as
\[
\left(s, a, s', r^g
\right), \forall \texttt{store\_id}, \texttt{run\_id},
\]
where $s$ and $s'$ denote the current and next states, $a$ is the selected action, 
and $r^g$ is the region-level reward. Since each store in each run contributes one 
data point, we construct transition tuples $(s, a, s')$ by pairing consecutive runs.

The system executes approximately three runs per minute, yielding
$24 \times 60 \times 3 = 4{,}320$
runs per day. With roughly $10$ stores per run, this corresponds to $43{,}200$ 
store-level data points per region per day. Across $4{,}000$ regions, full rollout 
would produce approximately
$4{,}000 \times 43{,}200 = 172.8 \text{ million}$
data points per day. Under the controlled $0.5\%$ rollout, only about $20$ regions 
are selected in each two-hour window, resulting in approximately
$864{,}000$ data points per day. Therefore, each 
two-day iteration produces approximately $1.73$ million transition samples, and the two iterations together yield approximately $3.46$ million samples.

\subsection{Policy Model and Parameters}
We parameterize the policy as a lightweight neural network that maps each store-level state to a discrete action. 
The input state is a three-dimensional feature vector $s \in \mathbb{R}^3$, and the action space contains 
$|\mathcal{A}|=5$ candidate actions. The policy network is implemented as a two-layer multilayer perceptron 
with hidden dimension $16$ and ReLU activations:
\[
Q_\theta(s)
=
\mathrm{Linear}(3,16)
\rightarrow
\mathrm{ReLU}
\rightarrow
\mathrm{Linear}(16,16)
\rightarrow
\mathrm{ReLU}
\rightarrow
\mathrm{Linear}(16,|\mathcal{A}|).
\]
At serving time, the policy selects the action with the largest predicted score:
\[
\pi(s) = \arg\max_{a \in \mathcal{A}} Q_\theta(s)_a .
\]
The model is trained offline using logged transition tuples $(s,a,s',r)$ collected from the controlled rollout. 
Training is initialized from a previously trained checkpoint and continued for $30$ epochs with mini-batches of 
size $32$. We use the Adam optimizer with learning rate $10^{-3}$, discount factor $\gamma=0.99$, gradient clipping 
with maximum norm $10$, and a target network that is refreshed every $2$ epochs. The final checkpoint is exported 
for offline evaluation and deployment.

\subsection{Training Curves}
\label{app:training-curves}

\begin{figure}[h]
    \centering
    \includegraphics[width=0.5\linewidth]{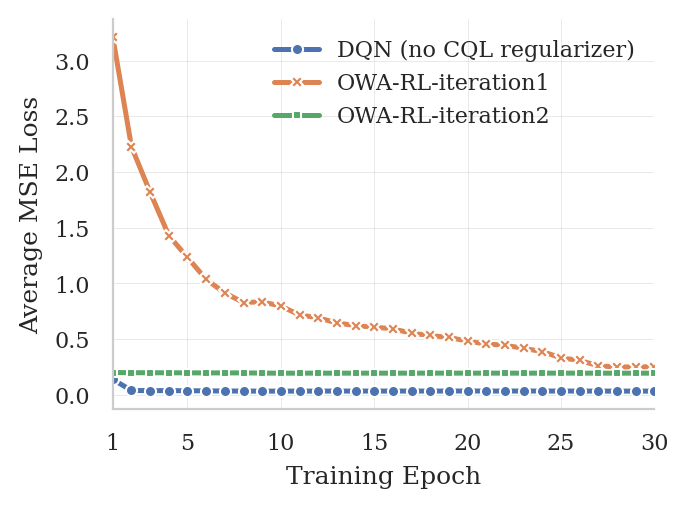}
    \caption{
    Offline training MSE loss across epochs. The DQN baseline without CQL regularization minimizes MSE faster, while OWA-RL maintains a higher loss due to the conservative penalty used to reduce unsupported-action overestimation.
    }
    \label{fig:training-curves}
\end{figure}

Figure~\ref{fig:training-curves} shows offline training MSE loss across epochs for the OWA-RL learner (two iterations) and a DQN baseline without CQL regularization. The DQN baseline reaches a lower MSE more quickly, while OWA-RL maintains a higher training loss because the conservative regularizer penalizes high values on unsupported actions. This behavior is expected: the goal of the conservative objective is not to minimize Bellman error alone, but to improve offline deployment stability by discouraging overestimated values for actions weakly supported by logged data.

\newpage
\section{Policy Behavior Under Reward Reweighting}
\label{app:reward-reweighting}

\begin{figure}[h]
    \centering
    \includegraphics[width=0.5\linewidth]{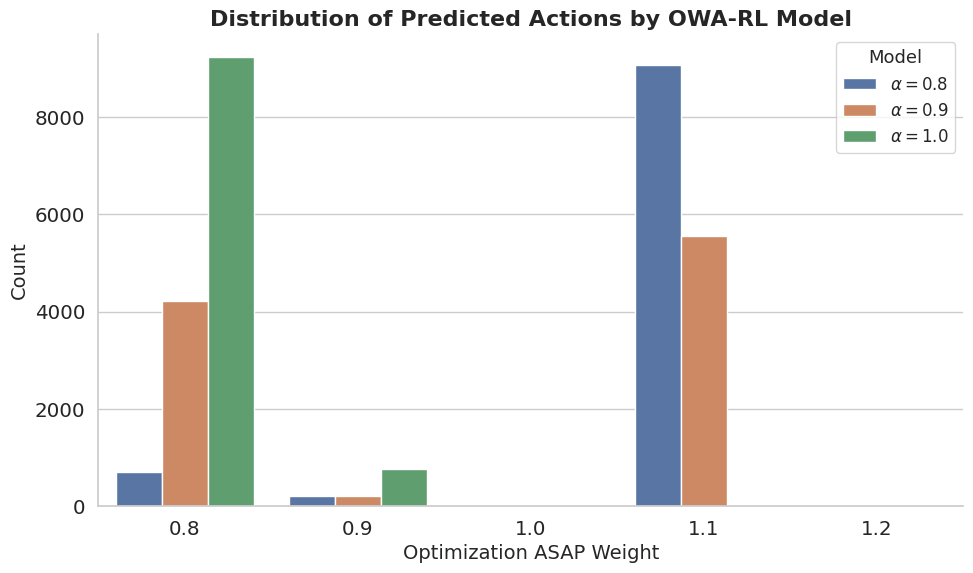}
    \caption{
    Predicted action distributions under different reward-weight settings. Changing the reward weights shifts the selected objective-weight multipliers, showing that the learned policy adapts its behavior to the desired speed-efficiency tradeoff.
    }
    \label{fig:action-dist}
\end{figure}

Before deployment, we evaluate whether the learned policy responds directionally to reward design. Figure~\ref{fig:action-dist} shows predicted action distributions under different reward-weight settings. Increasing the efficiency weight shifts probability mass toward lower ASAP-weight multipliers, making batch-compatible assignments more attractive to the optimizer; increasing the speed weight shifts mass toward higher multipliers, increasing the penalty on delay. This sensitivity check suggests that the policy uses the constrained action space to express the intended speed-efficiency tradeoff rather than collapsing to a single static action.

\newpage
\section{Examples of Drift-Monitoring Diagnostics}
\label{app:drift-monitoring}

\begin{figure}[h]
    \centering
    \includegraphics[width=0.9\linewidth]{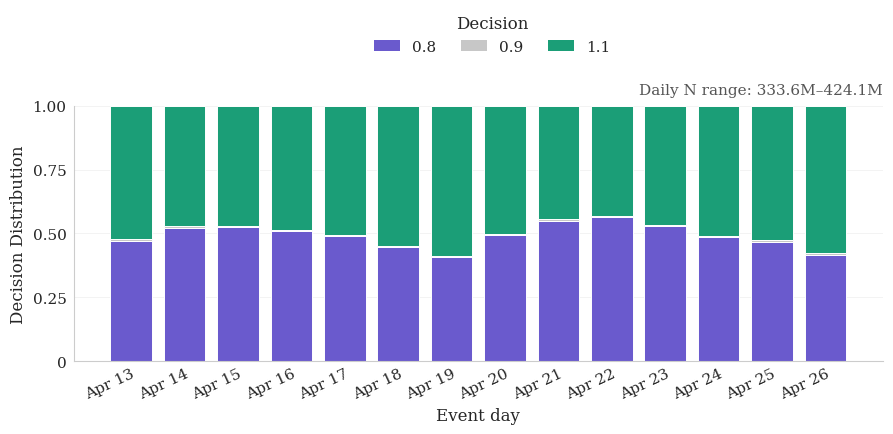}
    % \caption{
    % Daily distribution of selected ASAP-weight multipliers. Monitoring action distributions helps detect policy-behavior drift after deployment.
    % }
    \caption{
Daily distribution of selected ASAP-weight multipliers in production. 
Monitoring action distributions helps detect policy-behavior drift after deployment.
}
    \label{fig:action-trend}
\end{figure}

\begin{figure}[h]
    \centering
    \includegraphics[width=0.9\linewidth]{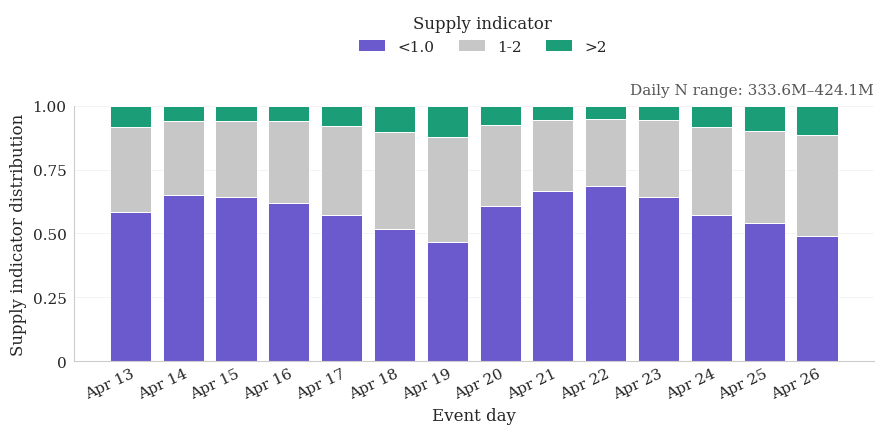}
    % \caption{
    % Daily distribution of the supply pressure indicator. Monitoring state-feature distributions helps detect marketplace drift that may affect policy reliability.
    % }
    \caption{
Daily distribution of the supply-pressure indicator in production. 
Monitoring state-feature distributions helps detect marketplace drift that may affect policy reliability.
}
    \label{fig:supply-trend}
\end{figure}

Figure~\ref{fig:action-trend} shows the daily distribution of selected ASAP-weight multipliers, while Figure~\ref{fig:supply-trend} shows the daily distribution of the supply indicator over the same period. These diagnostics are intended for production monitoring: large shifts in action distributions may indicate policy-behavior drift, while shifts in state-feature distributions may indicate marketplace drift that can affect policy reliability.

\end{document}